\documentclass[journal]{IEEEtran}
\usepackage{graphicx}          

\usepackage{tabularx}
\usepackage{booktabs}
\usepackage{multirow}
\usepackage{adjustbox}
\usepackage{url}
\ifCLASSINFOpdf
\else
\fi
\begin{document}
%
\title{Personalizing Education through an Adaptive LMS with Integrated LLMs}
%
%
%

\author{Kyle~Spriggs,
        Meng~Cheng~Lau,
        and~Kalpdrum~Passi
\thanks{K. Spriggs, M. C. Lau, and K. Passi are with the Bharti School of 
Engineering and Computer Science,  Laurentian University, Sudbury,
ON, P3E 2C6 CANADA e-mail: kspriggs@laurentian.ca, mclau@laurentian.ca, and kpassi@laurentian.ca.}}
\maketitle

\thispagestyle{empty}

\begin{abstract}
The widespread adoption of large language models (LLMs) marks a transformative era in technology, especially within the educational sector. This paper explores the integration of LLMs within learning management systems (LMSs) to develop an adaptive learning management system (ALMS) personalized for individual learners across various educational stages. Traditional LMSs, while facilitating the distribution of educational materials, fall short in addressing the nuanced needs of diverse student populations, particularly in settings with limited instructor availability. Our proposed system leverages the flexibility of AI to provide a customizable learning environment that adjusts to each user's evolving needs. By integrating a suite of general-purpose and domain-specific LLMs, this system aims to minimize common issues such as factual inaccuracies and outdated information, characteristic of general LLMs like OpenAI's ChatGPT. This paper details the development of an ALMS that not only addresses privacy concerns and the limitations of existing educational tools but also enhances the learning experience by maintaining engagement through personalized educational content.
\end{abstract}


%
\IEEEpeerreviewmaketitle

\section{Introduction}
%
%
%
%

 

\IEEEPARstart{W}{idespread} adoption of large language models (LLMs) is a recent phenomenon. Previous technologies lack the powerful automation utility LLMs provide. Simple wrappers over LLMs are novel but already commonplace. Unique innovations therefore require the combination of multiple technologies. In education, institutions distribute course materials and analyze student performance using digital learning management systems (LMSs)~\cite{turnbull2020learning}. Course-specific SaaS platforms also assist in the learning process~\cite{taufiq2021software}. In K-12, teacher attention is spread too thin to meet every student's needs; a problem worsened in post-secondary where self-study dominates~\cite{Koc2015}. Passive learning, such as rereading textbooks, is less effective than active methods~\cite{Rodriguez2018}. Standard LMSs rely on passive content, making the status quo sub-optimal. SaaS platforms, though helpful, are expensive and suffer from the same limitations. One-on-one instruction is the pinnacle of educational methods~\cite{Woolf2008}, as the learning experience can be personalized to actively moderate a student's cognitive load~\cite{Csikszentmihalyi1975}. Achieving this balance is hard, even for trained educators, and nearly impossible with a generalized approach~\cite{Wimpenny2013}. An optimal solution should adapt to the student, both during lessons and throughout a course.

We envision an adaptive learning management system personalized for each user. An AI agent can customize the experience via consistent user profiling. Such a system could overshadow conventional solutions. Traditional LMSs serve as delivery methods for academic materials, acting as proxies between teachers and students, allowing educators to manage content for student consumption, collect completed work, track academic progress, and communicate~\cite{turnbull2020learning}. This framework can be augmented to provide adaptive content tailored to users' goals. LLMs such as OpenAI's ChatGPT could be pivotal in such a system, but face issues like hallucinations, privacy concerns, knowledge gaps, and high costs~\cite{yao2024llm}. This paper discusses a prototype designed to address these issues. It integrates custom-trained, domain-specific LLMs for various tasks, balancing privacy and cost by choosing when to use secure models or third-party APIs. By extending models with specialized datasets and using retrieval augmented generation (RAG), the system reduces hallucinations and outdated information.

Development was divided into three phases. In Phase I, command-line scripts were designed for tasks like OCR and data management, and a Django web backend and React frontend were built. The system was then refined and containerized using Docker. Phase II integrated LLMs, developing a command-line interface with ChatGPT’s API and testing various configurations with RAG and vector embedding. Phase III compared the performance of various LLMs against human test scores in mathematics, reading, writing, reasoning, and coding. Data on resource utilization and performance was analyzed to benchmark the LLMs. This research and development process sought to evaluate the utility of expert systems and LLMs in an adaptive learning management system.

The remainder of this paper is structured as follows: Section~\ref{sec:litRev} provides a literature review, examining related work and foundational concepts. Section~\ref{sec:methodology}  details the implementation of the system, including the architecture and technologies used. Section~\ref{sec:experimentalDesign} describes the experimental design, outlining the methodologies and procedures followed. Section~\ref{sec:results} presents the results and discussion, analyzing the findings and their implications. Finally, Section~\ref{sec:conclusionAndFuture} concludes the paper and discusses future work, suggesting directions for further research and development.

\section{Literature Review} \label{sec:litRev}

\subsection{Artificial Intelligence in Education} \label{subsec:litRev/AI_inEdu}
Studies dating back decades before the advent of generative AI indicate that intelligent tutoring systems have a positive impact on student learning~\cite{SteenbergenHu2014}, in some cases comparable to that of human tutors~\cite{VanLehn2011tutor}. Now that AI has begun a societal boom, implementations using the technology to assist in the classroom are becoming increasingly popular. Current applications have been shown to benefit both students and teachers. Many of these systems are built around the concept of formative assessment, where evidence of students' learning is continuously logged and subsequently used to guide instructional adjustments and provide actionable feedback~\cite{McMillan2023formative}.

Student-oriented personalized learning systems may include additional features including generation of exercises, recommendation systems, and adaptive conversation~\cite{Zhang2021review}\cite{Wang2023review}. Some systems can detect how individual students learn best, helping them make informed learning choices, increasing student engagement, and lending to the generation of adaptive learning materials~\cite{Cutumisu2019}. Applications oriented towards teachers contain features like automated curriculum design, performance prediction for flagging struggling students, AI-generated lesson plans, frameworks for assessment, statistics tracking and analytics, generation of homework and test questions, and adaptive professional development systems~\cite{Martin2024review}. Technologies used to make these features possible consist primarily of machine learning, expert systems, and natural language processing~\cite{Zhang2021review}. AI tutoring systems can provide noticeable improvements to the quality of education across many domains and institutions, including K-12, post-secondary, and in the workplace~\cite{Wang2023review}\cite{Maity2019}. Trials have demonstrated tangible benefits such as increased student participation, increased learning performance, and reduced teacher workload~\cite{Wang2023review}.

Modern AI learning systems contain a combination of standout capabilities and debilitating shortfalls. Due to their natural language processing (NLP) backbone, LLMs have a proven track record of strong reading and writing skills, as well as reasonable competence in answering general knowledge and science-based questions~\cite{cai2024llm}. The same core building blocks provide AI learning systems relying on LLMs with competence in language-to-language translation, validating their inclusion in foreign language classes~\cite{kocmi2023translation}. Some prototype systems are designed to track user performance or categorize their preferences, and then use these cues to adapt the system's behaviour to match the user~\cite{Kim2020prototype}. LLM-based systems are reported to be capable of producing small blocks of code at or above the skill level of a post-secondary CS1 course. Test groups remark on the learning advantages of generating code samples, tutoring advice from chatbots during assignments, and LLM analysis of erroneous code~\cite{Manley2024CS1}.

Despite their promise, educational AI systems have many shortfalls. LLM-based systems are reputed to lack competency in both problem solving and tutoring on the topic of mathematics~\cite{hendrycks2021math}. The performance of LLMs also varies widely depending on the wording of semantically similar prompts, independent of the problem type~\cite{mizrahi2024llm}. Furthermore, LLMs are plagued by over-confident generation of inaccurate or outdated information, known as hallucinations~\cite{Jiang2023} (see section~\ref{subsec:litRev/llms}). Some prototype systems have had success avoiding hallucinations by using RAG and vector embedding, although this may require preventing LLMs from utilizing their training data when generating responses~\cite{dong2024prototype}. LLMs are also held back by either the financial cost of API calls to proprietary models, or in the case of self-hosted models, the requirement of high-spec hardware for LLMs with large parameter sizes. Finally, users with privacy concerns are rightfully cautious of using proprietary LLMs on account of how prompts and data fed into such models are often used to train future versions, allowing for the possibility that segments of one user's data may be relayed to another user, or stored on a company's server where it could be vulnerable to a data breach~\cite{Yao2024privacy}.

\subsection{Expert Systems} \label{subsec:litRev/es}
An expert system is a digital knowledge storage and retrieval system intended to assist in the resolution of problems using expert knowledge embedded into the system. They consist of a data pipeline that mirrors some aspects of the fundamental data science process~\cite{waterman1985expert}. First, raw data is collected from a reputable source, typically a human expert with a comprehensive understanding of the fine details of a particular domain. This data is initially stored in some format, such as a database, a collection of physical documents, or audio recordings. The information from this intermediate source is then codified and processed into practical knowledge that is relevant to a business or system requirement, or some general use case. Once translated into this more useful format, the knowledge is stored in a digital knowledge base. Rules for the retrieval of the stored knowledge are then established and programmed into the system, allowing end-users who lack relevant knowledge for a given problem to interact with the system and retrieve the knowledge that is specific to their issue~\cite{shuhsienliao2005es}\cite{HayesRoth1984}.

\subsection{Large Language Models} \label{subsec:litRev/llms}
Large language models (LLMs) are a branch of artificial intelligence (AI) based on transformer architecture~\cite{singh2021NLP}. We will build an understanding of the fundamental nature and resulting characteristics of LLMs by beginning with their ``atomic,'' low-level conceptual precursors, and progressively scaling back towards more macroscopic, corollary technical concepts. Deep learning is a machine learning (ML) process that makes use of neural networks~\cite{LeCun2015Deep}. Their function is beyond the scope we will explore. Transformers are ML models trained using deep learning to perform specialized tasks like computer vision or natural language processing (NLP)~\cite{Lin2022Transformers}. LLMs are a type of transformer that is trained for NLP tasks, making them specialized for text-completion and question-answering tasks~\cite{Singh2021}. This requires an extensive training process. Enormous quantities of sample text are passed into the LLM as input, and parameter settings are adjusted to produce consistently more desirable output~\cite{shanahan2023llms}.

Training of transformers, LLMs, and other AI technologies is performed using a combination of several ML techniques, including supervised and unsupervised learning. Supervised learning involves the input of large quantities of labelled datasets to train algorithms to accurately classify data or predict the next output in a sequence~\cite{Singh2021}. In terms of NLP models, this entails predicting the most suitable word or phrase in a partially completed sentence, paragraph, or larger body of text. Unsupervised learning is characterized by training where no labelled data is provided to the system. Transformers and their LLM counterparts are trained using some subset of these techniques depending on the goals of the researcher\cite{zhao_survey_2023}. Large language models can be optimized using a process known as fine-tuning. This involves achieving desired outputs by retraining models on specific examples, with input/output pairs that demonstrate the intended behaviour. Fine-tuning of general-purpose base models, which were pre-trained exclusively using the above ML techniques, can rapidly modify their outputs to fit a desired use case~\cite{zhao_survey_2023}. LLMs can be merged, combining the strengths of each precursor model. Many open-source LLM models released into the public domain use these strategies to improve performance benchmarks of base models with a comparatively small amount of effort, time, and computational power. Once training is completed, it is possible to augment the behaviour of LLMs using techniques such as vector embedding, retrieval augmented generation (RAG), and prompt engineering~\cite{Singh2021}\cite{zhao_survey_2023}.

Vector embedding involves the conversion of words or phrases into sequences of weights that represent the string's semantic similarity to other words or phrases. This series of weights is called a vector. Vectors can be stored in format-specific databases called \textit{vectorstores} for subsequent retrieval. In this way, the semantic meaning of a string can be stored numerically, and later cross-referenced with other vectors to determine how closely related the strings are semantically~\cite{Heimerl2018}. This is the basis of semantic (vector) search, as contrasted with conventional text-based (keyword matching) search. Retrieval augmented generation allows for complete documents, such as text files, to be parsed and uploaded into a local \textit{vectorstore}, with a large enough chunk size that entire paragraphs may be stored as individual vectors~\cite{Li2022}. Using RAG allows the user’s local \textit{vectorstore} to be accessed before retrieving output from the LLM’s core response pathway, i.e. bypassing the retrieval of data stored in its neural network~\cite{Li2022}. This can be a valuable way to improve certainty in the factual validity of an LLM's output~\cite{Jiang2023}. Outputs are otherwise uncertain due to the prevalence of hallucinations, which are inaccurate responses produced by LLMs due to them being exposed to misinformation or conflicting data during training~\cite{Yao2023}. It is theorized that due to the nature of how they are trained, a process that is integral to their design, hallucinations are an innate and unavoidable pitfall of LLMs~\cite{Yao2023}\cite{Xu2024}.

LLMs can be accessed via an API from proprietary sources such as OpenAI for a fee. Alternatively, it is possible to host open-source or custom-trained models locally, substituting financial cost for hardware resource utilization. Table~\ref{tab:llm_size} summarizes the approximate hardware requirements of various LLM parameter size configurations for stable and efficient operation.

\begin{table}[]
    \caption{Requirements by Parameter Count}
    \centering
    \begin{tabular}{l|l}
    \hline
         Parameter Count & Memory Demand: \\
         \hline
         7B & 8GB RAM \\
         13B & 16GB RAM \\
         70B & 64GB RAM \\ 
         \hline
    \end{tabular}
    \label{tab:llm_size}
\end{table}

\section{Methodology} \label{sec:methodology}
The development methodologies of \textit{DevOps} were adopted very early in the initial research and planning stage of the project, and referenced as an overarching philosophy throughout the development process. Short for ``development and operations,'' \textit{DevOps} builds on many principles used in Agile project management, breaking away from the rigidity and bureaucratic ideas that hamper progress within the waterfall model~\cite{Jabbari2016}.

\textit{DevOps} accelerates development through strategies like continuous integration and continuous delivery (CI/CD). In continuous integration, a version control system is utilized to merge code changes made by each developer into a central repository, where they can be tested immediately after their initial design, and integrated into the main project branch shortly thereafter. This allows changes to become functional parts of the project without the delay of waiting for the next development stage, as in the waterfall model. This strategy also highlights the \textit{DevOps} principle of communication and collaboration, by uniting the functions of developers and operations into a streamlined, integrated process~\cite{Jabbari2016}. Git version control software and GitHub repositories were used during this project to achieve continuous integration. 

In continuous delivery, code changes are built, tested, and prepared for deployment shortly after they are designed, often be the same developer who drafted the changes. This creates a trend towards rapid delivery by allowing bugs to be identified immediately after they are introduced. The \textit{DevOps} principle of reliance on automation also comes into play here, using automated testing tools to speed up the process~\cite{Jabbari2016}. We employed these practices by testing each functional unit of code before moving on to subsequent sections, often on the same day that they were produced.

Automation tools are used in \textit{DevOps} throughout the development process, contributing to rapid deployment goals. In this project, this took the form of frameworks like Django, which accelerated the design and maintenance of the web backend. \textit{DevOps} also takes the stance that we should begin with the end goal in mind, and consistently work towards that goal. This entails an emphasis on modular design, implying encapsulation of processes so that all aspects of the project can be scaled as necessary~\cite{Jabbari2016}. We used a modular approach in our design, encapsulating sections of the project using Docker containerization and virtual environments.

\section{Implementation} \label{sec:implementation}

\subsection{Expert System} \label{subsec:imp/es}
In Phase I of our implementation, a prototype expert system was developed using a combination of Python, JavaScript/TypeScript, JSX/HTML, CSS, Shell, and SQL programming languages. Functioning as a web app using a Django backend and React frontend, this system combines multiple functions involving OCR, text-based search, and HTML scraping. At present the system effectively solves user questions  by matching their query with the four closest options loaded into the system’s knowledge base.

\begin{figure} [ht]
    \centering
    \includegraphics[width=1\linewidth]{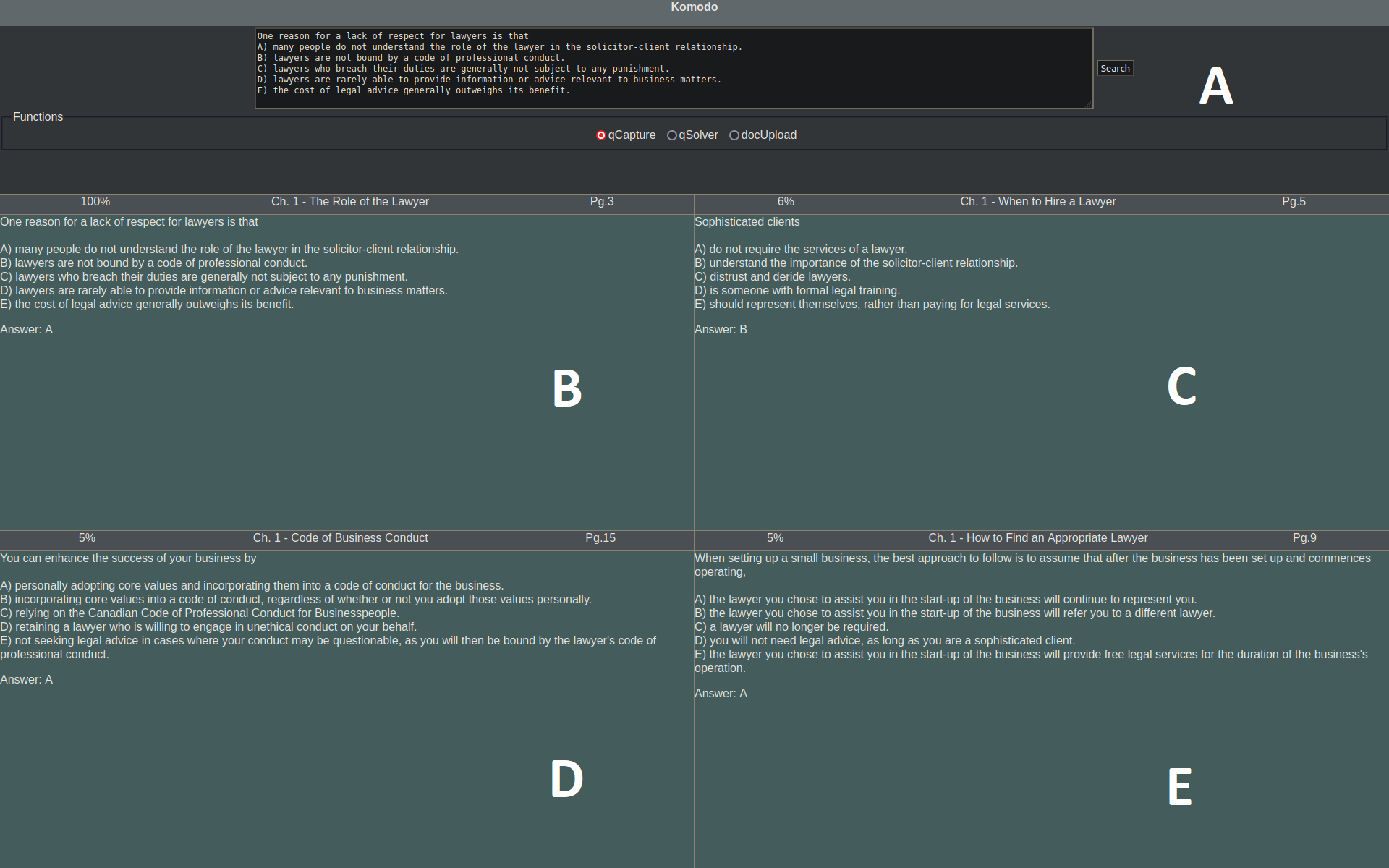}
    \caption{Screen capture of user interface. A:~Query input, B:~ContentPanel-1 (closest match), C:~ContentPanel-2, D:~ContentPanel-3, E:~ContentPanel-4.}
    \label{fig:UI_1}
\end{figure}

Fig.~\ref{fig:UI_1} showcases the UI design developed during Phase I. Three functions are included: \textit{qCapture} (question capture), \textit{qSolver} (question solver), and \textit{docUpload} (document upload). The \textit{qCapture} function is intended to assist the user in using \textit{qSolver} on an existing problem. The user is prompted to draw a box with their cursor. A screenshot is captured and cropped within the bounds of the user-defined box. The image is scanned using an OCR model, and the resulting text is passed into the UI’s search field. The \textit{qSolver} feature compares the search text to each question stored in the test bank database. Matches are based on the similarity of word composition. The closest matches are passed into the content panel, one per tile, along with any relevant accompanying data (i.e. answer, topic, chapter, and page number). The \textit{docUpload} feature currently has no UI, and exists as a series of command line scripts. It allows test bank PDFs to be converted into formatted text using OCR, which is broken down into individual questions, parsed into sub-fields, and saved into a MySQL database. 

Collectively, these features allow a user to upload a pre-configured test bank containing questions with complete coverage of the related textbook's key concepts, capture a screenshot of a related question that they are struggling with, and match their problem with the four most similar questions existing within their database. The matching questions include an answer, a short description of the related topic or concept as listed in their textbook, and the page number within the textbook where that concept is explained in full.

\begin{figure} [ht]
    \centering
    \includegraphics[width=1\linewidth]{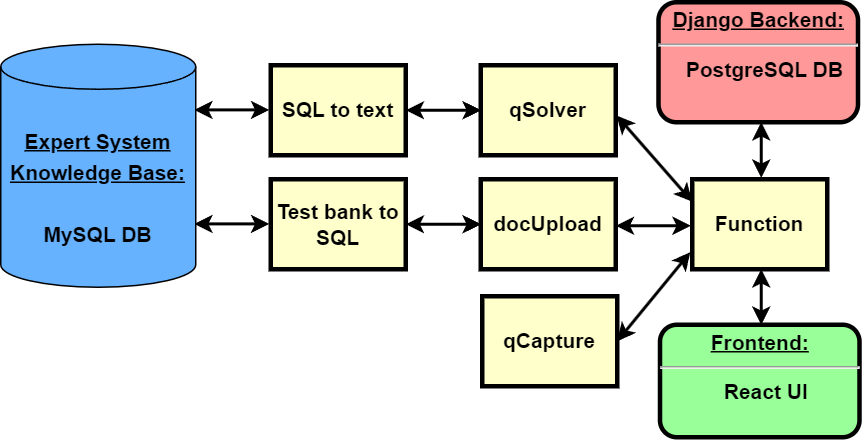}
    \caption{Backend DFD}
    \label{fig:backendDFD}
\end{figure}

\begin{figure} [ht]
    \centering
    \includegraphics[width=0.7\linewidth]{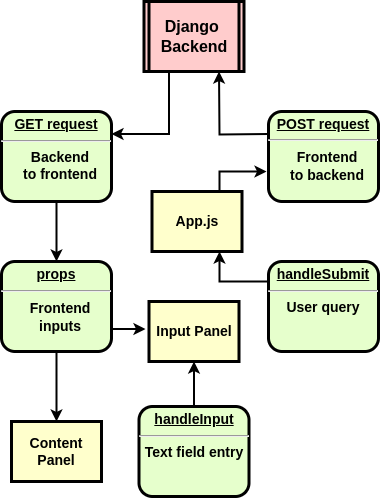}
    \caption{Frontend state diagram}
    \label{fig:frontendDiagram}
\end{figure}

While the backend stores data within a PostgreSQL database, the frontend maintains a much smaller sample of relevant data. Fig.~\ref{fig:frontendDiagram} depicts the core mechanism responsible for triggering the transmission of this data, any time user input is received from event listeners built into the UI.

In \textit{React}, components are stored in browser memory. Since browsers restrict their allocation of memory, web applications must limit the amount of data that is rendered into the DOM. This also decreases loading times and increases responsiveness. There are two solutions to this problem. One comes in the form of \textit{React}’s virtual DOM. Any data that is programmed to be rendered into a React webpage is translated into a virtual DOM element. On the first load, React renders all of this data. On subsequent loads, the virtual DOM is compared to the browser’s actual DOM, and only missing elements are updated. The second solution is to simply limit the amount of data that is passed into the virtual DOM for any given page. This involves the curation of data elements stored in the backend, such that only the required parts of a dataset are passed into variables that will be rendered to the DOM.

This becomes possible through \textit{React}’s state management system. State may be handled globally, i.e. by defining a variable to track state in the application’s root component, or locally, i.e. using a dynamic variable encapsulated within an individual child component. React includes special methods called ``Hooks” which offer programmers an extended level of control over components without the need to manually implement workarounds. These include the \textit{setState()} and \textit{useState()} methods, which present themselves as a solution for managing component states.

Komodo uses these Hooks to define state at the global level, within \textit{App.js}, as well as locally within certain child components, such as the \textit{InputPanel} and \textit{ContentPanel}. Here, the global state keeps a record of all data fields that may be needed in any of its child components, which it passes into the required component when necessary via React props. Each child component’s local state is composed of a subset of the fields within the app’s global state, which is itself a subset of the data stored in \textit{Django}’s \textit{FrontendInputModel}.

When user input modifies the UI, the state of the corresponding component is updated using the \textit{handleInput()} method. Any time a child component needs to render data it does not yet have, it initiates a POST request in the root \textit{App.js }component by calling the \textit{handleSubmit() }function. This only happens when the user triggers the event by interacting with the UI, and so at present only the \textit{InputPanel} is equipped to handle it. After \textit{handleSubmit()} triggers the \textit{POST} request, the state of the \textit{InputPanel} is passed to the \textit{FrontendOutputModel }in the backend. This data includes search field submissions and function selection. Based on the data submitted, some app feature may be activated, and a \textit{GET} request may be chained into the event, causing specific data from the backend’s FrontendInput model to update both the global app state, as well as the state of any relevant child component via \textit{React} props.

\subsection{LLM Question-Answering Machine} \label{subsec:litRev/llm_qam}
Phase II of the project began with the implementation of a command-line interface written in Python that interacted with OpenAI's API, allowing for a single query to be relayed to ChatGPT-3.5 Turbo. This was expanded with a simple loop to provide an unlimited series of query-response cycles, along with basic commands to break the loop, adjust the temperature (randomness) of the AI's responses, limit the maximum number of tokens in the response, and switch to other OpenAI models. A Python list was added to store a log of messages between the user and agent throughout the conversation. By piping a set number of recent messages from the conversation history into the context of each subsequent prompt, the AI developed a basic form of short-term memory, able to respond with an understanding of the persistent topic.

This basic loop was re-created using the altered syntax of the \textit{LangChain} Python library. \textit{LangChain} facilitates the use of multiple LLM models from a single command, increasing testing speed and modularity. Within this framework, OpenAI's suite of LLMs were tested, including ChatGPT-3.5 Turbo, GPT-3.5 Turbo Instruct, and GPT-4. The ``instruct" version of GPT-3.5 was found to return semantically similar responses to the baseline version, but with less verbose output.

LangChain's capabilities were expanded with the integration of the Ollama extension. Ollama opened the door to experimentation with self-hosted LLMs, which are downloaded to the host machine and run locally, without API calls or bandwidth usage. Initial testing was done using Mistral-7B and Llama2-7B. The 7 billion parameter models were chosen due to their reduced system requirements compared to higher-performance but lower-speed versions of the same LLMs. See subsection~\ref{subsec:litRev/llms} for a table~\ref{tab:llm_size} summarizing hardware requirements corresponding to different parameter ranges.

Informal experimentation was performed on the self-hosted and proprietary models' capabilities in system prompt modification, RAG, vector embedding, and prompt engineering. System prompt modification involved context injection to augment the personality, speech style, and flavour of responses. Using internal system prompts like ``You are Luke Skywalker" established an identity for the agent, which it would emulate insofar as it had knowledge about that persona. In the above example, a user query of ``Who's your daddy?" would return a response similar to ``My father is Darth Vader, otherwise known as Anakin Skywalker."

For RAG, two directions were investigated. First, RAG from text documents were tested. Here we used output from PDF transcriptions obtained from the Phase I system as an input for the LangChain pipeline. The text loader module parsed the .txt input files, split the document into overlapping chunks, translated each chunk's semantic data into a vector, and stored the collection of vectors into a FAISS vectorstore in memory. Embedding models tested during the intermediate stage of this process include OpenAI's Ada-002, SFR-Embedding-Mistral, voyage-lite-02-instruct, and mxbai-embed-large-v1. Ada-002 was used for its speed and consistency in all subsequent tests.

Next, we explored RAG from webpage URLs. In these tests, we used the same webpage scraping library as in Phase I; \textit{BeautifulSoup}. By inputting a URL into the appropriate field, the scraper extracted all text on the page, and passed it into LangChain's pipeline, onward to the same RAG toolkit discussed above. In both the RAG from local document and RAG from URL tests, a consistent system prompt was used:

$`` `` ``Answer\;the\;following\;question\;based\; only\; on\; the\; \\ provided\; context:
\;\; \langle context \rangle \; \{context\} \; \langle /context\rangle
\\ \;Question: \{input\}"""$

User queries related to the parsed input source were answered with surprising accuracy. To test for hallucinations, queries unrelated to the source were used. In all cases, the system responded that the requested information was not present in the source articles, rather than misrepresenting falsehoods with unwavering confidence as was the case with some non-RAG tests. Some experimentation with prompt engineering was also performed. This proved valuable as a means of adjusting the AI's responses to specific questions. Lessons learned in these experiments were employed to great effect in Phase III of the project.

In terms of hardware bottlenecks, testing on multiple systems revealed RAM, CPU, and GPU quality to be major factors in execution time. Larger parameter-size models took upwards of four minutes to complete on lower-spec computers without GPU assistance. AI tutors typically use a single LLM as the core of their system, adding techniques like prompt engineering and fine-tuning to modify the LLM to behave appropriately. Data retrieved by the system often relies on training datasets as a source. RAG was shown to offer a superior alternative, allowing for the source to be curated for the specific use case, and for sources to be provided with the results.

Further optimization can be achieved by using multiple AI models depending on the sub-task. For example, Mistral for NLP, CodeLlama for programming, OpenAI's Ada as an encoding model, and Nougat as a vision transformer. This delegation of sub-tasks separates the data pipeline into pieces that can be managed with a higher degree of control. When the data pipeline is segmented further, it begins to resemble an expert system more than a general-purpose AI model. This revelation served as inspiration for Phase III of the project, where we investigated the relative strengths and weakness of individual LLMs (see sections~\ref{sec:experimentalDesign} and~\ref{sec:results}). Future versions of Komodo will make use of the research conducted here, substituting the text-based search function from Phase I with LLM-powered semantic search, populating a vectorstore with the contents of the expert system's knowledge base, and incorporating web scraping with RAG to expand the corpus of information accessible within the system using user-curated resources.

\section{Experimental Design} \label{sec:experimentalDesign}

\subsection{Benchmarking System Design} \label{subsec:benchmarkingDesign}
Phase III of the project consisted of an analysis of the strengths and weaknesses of a choice selection of LLMs. The intention was to discern which models could be relied on for a given task type, while acknowledging the associated costs in their operation, including financial costs and hardware resource usage. In addition to comparing LLMs to one another, the research aimed to clarify what the general capabilities of LLMs were, and how these capabilities fared against typical human performances in similar tests.

To achieve this, an extensive battery of tests was designed. Most question sets were based on standardized tests applied in schools across Canada, ranging from the K-12 level upwards to post-secondary. Five subject categories were tested: mathematics, reading, writing, reasoning, and coding. From each of these categories, a subset of three question sets were assembled according to estimations of their difficulty, from easy, medium, and hard, based on informal testing in Phase II of the project. For each of the three difficulties, ten questions were selected to be tested, except for the coding and writing categories, due to limitations on grading speed. The questions were distributed as shown in Table~\ref{tab:test_question}.

\begin{table}[]
    \caption{Test Question Distribution}
    \centering
    \begin{tabular}{l| c |c| c}
    \hline
         Category: &  Difficulty: & Questions: & Rounds: \\
         \hline
         Mathematics & E & 10 & 3 \\
         & M & 10 & 3 \\
         & H & 10 & 3 \\
         Reading & E & 10 & 3 \\
         & M & 10 & 3 \\
         & H & 10 & 3 \\
         Writing & E & 10 & 3 \\
         & M & 10 & 3 \\
         & H & 1 & 1 \\
         Reasoning & E & 10 & 3 \\
         & M & 10 & 3 \\
         & H & 10 & 3 \\
         Coding & E & 5 & 1 \\
         & M & 5 & 1 \\
         & H & 5 & 1 \\
         \hline
    \end{tabular}
    \label{tab:test_question}
\end{table}

Note the question count for the hard writing question and the three coding question sets. This is due to the time-intensive nature of manually grading these questions. All question sets denoted above with 10 questions were arranged in multiple-choice format, with an automatic grading system implemented using Python. Since LLM responses are somewhat unpredictable, with a tendency toward verbosity given their NLP underpinnings, consistent output of a single character response (i.e. A, B, C, D, or E) was not possible for most models. Therefore, a combination of filtering the AI responses using regular expressions, combined with manual grading for many questions, was necessary.

To gain the desired insight, ten LLM models were chosen from a stratified sample of over fifty candidates researched. Each model was tested on the entire battery three times over. As noted above, some questions were graded for only one of the three rounds. This made for a total of 3,780 data points that had to be carefully tested, filtered, verified, and graded. Table~\ref{tab:llm} lists the LLMs tested.

\begin{table}[]
    \caption{LLMs Tested}
    \centering
    \begin{tabular}{l |c |c}
    \hline
    LLM & Parameters: & HDD Space: \\
    \hline
    GPT3\_5 & 175B & - \\
    GPT3\_5\_I & 175B & - \\
    GPT4 & $\sim$1.7T & - \\
    LLAMA2\_7B & 7B & 3.8GB \\
    MISTRAL\_7B & 7B & 4.1GB \\
    GEMMA\_7B & 7B & 5.2GB \\
    FALCON\_7B & 7B & 4.2GB \\
    CODELLAMA & 7B & 3.8GB \\
    WIZARD\_MATH\_7B & 7B & 4.1GB \\
    PHI & 2.3B & 1.7GB \\
    \hline
    \end{tabular}
    \label{tab:llm}
\end{table}

The question sets themselves were either sourced directly from standardized tests or manually designed to match the intended difficulty and style. Each question was formatted into the appropriate JSON file, where it was read by the testing module one at a time. Results were stored in the local filesystem in batch sizes of 30 questions per file, to prevent data loss if a test was interrupted. This was a relevant factor since each LLM could take over an hour to complete its circuit. After testing, test results were re-formatted to extract specific fields, which were then converted into a Pandas DataFrame and exported to Excel. The Excel files could then be filtered, verified and graded. Details on the contents of the aptitude tests follow in sections~\ref{subsec:mathTestDesign} to~\ref{subsec:codingTestDesign}. In addition to LLM problem-solving aptitude, the system's resource usage was also tracked across a range of datapoints (see section~\ref{subsec:resourceTestDesign}).

\subsection{Mathematics Test Design} \label{subsec:mathTestDesign}

Informal tests performed in Phase II suggested that LLMs had significant difficulty with mathematics problems. In anticipation of this, question sets were chosen primarily from standardized elementary school tests, rather than including advanced problems from the post-secondary level. For easy difficulty, problems from the elementary school EQAO Grade 3 Mathematics test were selected. Practice problems from the EQAO Grade 6 Mathematics test were chosen for medium difficulty. For hard, the ACT Mathematics Test, representative of skills required for entrance into post-secondary institutions, was used as a source~\cite{act2024_1}\cite{eqao2012}\cite{eqao2023_2}. All of these sources included a mixture of text-based and diagram-based problems. Since the models to be tested are not capable of interpreting images, any questions containing diagrams were omitted. Ten questions were selected for each difficulty. All questions were multiple choice style.

\subsection{Reading Test Design} \label{subsec:readingTestDesign}
Tests performed in Phase II indicated that all models possessed very strong capabilities in answering reading questions, regardless of difficulty level. In light of this, attempts were made to source especially challenging problems, many of which required parsing through and interpreting lengthy passages. For easy difficulty, selections were taken from the reading section of the 2015 Ontario Secondary School Literacy Test (OSSLT). Medium-level questions were sourced from the ACT Reading test, and hard questions from practice problems at lsac.org for the LSAT Reading Comprehension test~\cite{act2024_2}\cite{eqao2015_2}\cite{LSAC2024_2}.

Ten multiple-choice questions were chosen for each difficulty. Questions from medium difficulty were typically between 700 and 800 words long; many hard-level problems required reading two separate passages and answering questions comparing the contents of each passage.

\subsection{Writing Test Design} \label{subsec:writingTestDesign}
In this category, a variety of question types was selected to investigate different dimensions of models' writing ability. For easy difficulty, ten short general knowledge questions were designed to test how accurately each LLM could follow simple writing instructions. These questions required a small amount of knowledge, which was expected to be easily accessible due to the breadth of training data each model was exposed to. A simple pass/fail marking scheme was devised to assess their performance.

For medium difficulty, ten multiple choice practice questions from the OSSLT Writing Test, which assesses knowledge of grammatical structure, semantics, and word choice were selected. Since practice question sets for these tests contain only a few multiple choice questions, problems from 2015, 2019, and 2023 were compiled to reach the desired count of ten questions~\cite{eqao2015_1}\cite{eqao2015_2}\cite{eqao2019}\cite{eqao2023_1}. 

The hard writing test was given special consideration. A second multiple-choice test was regarded as a poor means of assessing the true writing ability of the assembled LLMs. Instead, models were tasked with writing a short essay, sourced from the ACT Writing Test~\cite{act2024_3}. Several teachers volunteered to grade the essays; results from section~\ref{subsec:writingResults} represent the mean of their scoring.

The essay writing question involved reading a short passage, analyzing three perspectives related to the passage, and then writing an essay within the specifications provided. The instructions given to the LLM models were identical to those in the source material, with the addition of a 500-word limit to reduce marking time and add consistency between responses. Note that of the sample essays included in the source material, most samples averaged well below 500 words.

Grading for this topic mimicked the marking scheme used on the actual ACT. A rubric was carefully compiled based on the parameters established from the source test materials~\cite{act2024_3} and all marking was done according to these stipulations. Scoring was recorded in the same format as for human subjects taking the ACT writing test; grades (out of six marks each) were assigned independently across four performance categories: ideas and analysis, development and support, organization, and language use. This produced a total out of a maximum of 24 points. As per ACT guidelines, this total was then represented as a proportionate score out of 36 points, following the standard procedure outlined in the ACT marking instructions~\cite{act2016}, allowing scores to be compared to the other categories of the ACT.

The ACT’s College Readiness Benchmark Scores are empirically derived benchmarks, where students who meet the minimum benchmark score have a 50\% chance of achieving a B grade or higher in related college courses~\cite{act2023}. Writing scores are a part of the ELA score but do not impact English or Composite scores. The minimum benchmark score on this scale for ELA is 20. Since the writing test scores are used to calculate ELA totals, we will use a cut-off of 20/36 in our final pass/fail assessment of each LLM’s writing ability, to establish which LLMs are at or above the minimum viable level for post-secondary success according to the College Readiness Benchmark scale. 

One essay was produced by each of the ten LLM models, and conscientiously graded according to the rubric. The names of the LLM models that produced each essay were omitted from the copies provided to the graders, and organized in random order, to eliminate marker bias.

\subsection{Reasoning Test Design} \label{subsec:reasoningTestDesign}
Informal tests in Phase II of the project suggested a significantly stronger ability in verbal reasoning than with problems involving arithmetic and mathematics. Based on these results, it was hypothesized that this trend would carry over to testing in Phase III. As such, verbal classification questions were selected for the easy difficulty of this category, and numerical reasoning problems were sourced for the medium difficulty level. Ten questions were selected for each of these difficulties, from a test bank of interview practice questions~\cite{indiabix_1}\cite{indiabix_2}. All questions were in multiple-choice format.

For the hard difficulty in this category, ten questions from the Law School Admission Test (LSAT) pertaining to logical reasoning were selected. This test is known for being very difficult for human subjects, and so was assumed to be especially challenging for LLMs as well. The question set used was obtained from online practice questions at lsac.org; since they are intended to prepare law students for the entry exam, these problems are unwavering in difficulty. A consistent string containing the directions provided on the source webpage was prepended to the query for each question~\cite{LSAC2024_1}.

\subsection{Coding Test Design} \label{subsec:codingTestDesign}
For this category, multiple sources were consulted to produce five custom problems per category. All problems were based on searching algorithms, for consistency~\cite{clrs2001book}\cite{gg2024web}\cite{w3_2024web}. Each difficulty level was made incrementally more challenging by requiring a minimum solution corresponding to progressively higher time complexities. Within easy difficulty, most problems had time complexities of O(n) or O(n log n). For medium and hard, either time complexities were greater, or similar problems were reputed to be suitably difficult for human subjects. All questions specified a solution to be written in the Python programming language. 

For all problems, the AI models were instructed to write a single Python function named \textit{test()} with specific input and return types. This was intended to provide consistency, make testing more empirical, and accelerate the grading process. Rather than including ten questions per difficulty, as in most other categories, only five per difficulty were included here due to the time required to design and test each problem.

Special considerations were needed in the grading of this problem type. Analysis of style was beyond the scope of these tests. To ensure consistent, empirical results were gathered, unit tests were written for each problem, outputting the result as a binary pass/fail. The appropriate portion of the AI's response was manually copied into the matching unit testing module, and results were logged for later analysis. Given five problems per difficulty, across three difficulties, with ten LLMs, this amounted to: 5 * 3 * 10 = 150 coding problems that had to be manually graded. Note that the questions selected can be considered to be of easy, medium, and hard difficulty for beginner programmers. This design choice was intended to mirror the type of questions encountered in a post-secondary CS1 course, to compare the ability of LLMs against students at this skill level.

\subsection{Resource Utilization Test Design}~\label{subsec:resourceTestDesign}
In addition to benchmarking of problem-solving abilities, the LLMs' system resource usage was recorded across a range of metrics, for each question solved. These records were averaged across all attempts at a given question and then arranged into tables and charts for analysis. Collecting this data was considered valuable in gaining an understanding of the relative system demands across each LLM tested, and to determine how computationally intensive each question in our test bank was to solve. Such readings can also be useful in identifying aberrant behavior such as memory leaks. Table~\ref{tab:hardware_met} below lists tested metrics, along with an explanation of what they represent.

\begin{table}[]
    \caption{Hardware Metrics Tested}
    \centering
    \begin{tabularx}{\columnwidth}{l|X}
         \hline
         Metric & Explanation \\
         \hline
         Avg CPU\% & Average \% of total system CPU in use during execution. \\
         Avg Mem\% & Average \% of total system RAM in use during execution. \\
         System Time & Time process spent occupying CPU in kernel mode (direct access to hardware). \\
         User Time & Time process spent occupying CPU in user mode (indirect access to hardware). \\
         Execution Time & Total runtime of process tree. \\
         \hline
    \end{tabularx}
    \label{tab:hardware_met}
\end{table}

Many strategies were attempted to achieve this detailed level of system data collection. Initially, Celery was used as a task queue. Celery's broker sends and receives messages to the system kernel, managing thread usage and launching tasks as subprocesses. \textit{RabbitMQ} was used as a broker over Redis and other options, due to its more robust and stable nature. A \textit{SQLite3} database was employed as a backend for passing results from celery workers back to the test manager. Flower, a Celery worker monitoring agent, was set up to collect the desired resource metrics. However, Flower cannot access all of the above usage details. Furthermore, Celery disallows the creation of subprocesses via Python's \textit{subprocess} library, which was integral to the testing module. Thus, Celery and Flower were abandoned.

Next, control groups were investigated as a potential solution. These are used in systemd-based operating systems. \textit{Systemd} is the first process that runs after boot on post-\textit{SystemV }UNIX systems. It manages all other services and processes on the system and uses control groups (\textit{cgroups}) to manage processes, track their resource utilization, and set limits on allowed usage of system hardware components. Since \textit{cgroups} isolate parent processes and any subprocess hierarchy spawned by them from the rest of the system, they are ideal for accurately identifying an application's resource draw.

In this project, after excluding Celery and Flower as options, the intention was to manually create a \textit{cgroup} for each test instance, which would then encapsulate all related subprocesses. The \textit{cgroup} could then be accessed at the end of each test to return readings of total and mean usage across the entire process hierarchy during its lifespan. 

Due to difficulties in directly creating and accessing \textit{cgroups} using low-level \textit{systemd} commands, the power of \textit{cgroups} was exploited by accessing groups created automatically by the kernel upon the initialization of our testing application using the \textit{getrusage()} function from Python's ``resource" library. This function is capable of accessing \textit{cgroup} usage data of any input process identifier (PID). We made use of this in order to retrieve the system time and user time of our application's execution. Because \textit{getrusage()} provides results from the queried PID and its child process hierarchy separately, these values were summed to find the total usage of the testing application.

However, the LLMs tested were not considered to be subprocesses of the testing application's PID in terms of \textit{cgroup} encapsulation. This is because Ollama runs as a backgrounded daemon between executions of the locally hosted LLMs it serves, even after the parent testing application process is terminated. Thus, a workaround had to be devised to collect data from the local Ollama server's usage, unique to each separate question tested. The solution implemented included the identification of the Ollama server's PID, termination of the process before each test, verification of the process's termination, initialization of a new Ollama server instance, and then access to the new Ollama server's usage data after each test was completed. This method granted access to the CPU system time and user time across all associated processes during each separate test.

While the above method could be used to monitor some hardware usage metrics, we also desired access to the average CPU percentage, average virtual memory percentage, and overall execution times for each question tested. Therefore, a separate method was designed to collect these data points, using the \textit{psutil} (Python system and process utilities) library. This library is capable of returning system-wide CPU and virtual memory percentages. Since \textit{psutil} does not provide average readings over time like \textit{getrusage()}, readings were collected at intervals of 0.25 seconds during each test. Averages were computed for CPU and memory usage. Execution time was found by recording the start and end time of each test using the /usr/bin/time command, and computing the difference. Data points across all of these variables were exported to a file along with aptitude results for each question.

\subsection{Data Representation} \label{subsec:dataRepresentation}
To represent all accumulated data in a useful format for analysis, data science principles were applied. Each question first had to be graded, before any statistics could be evaluated. This was accomplished through the creation of a custom Python grading module, involving the filtering of AI responses using regular expressions to detect whether they included the character denoted in the solution column for any given question (i.e. A, B, C, D, or E). This allowed large portions of the raw data to be automatically graded in a binary pass/fail format. Inconclusive results were later analyzed and graded manually.

All data points were then merged into a single frame, flattening the representation into a single table with 3,780 rows and 19 columns, in first normal form. This was accomplished by parsing intermediate test output stored in a series of raw text files into multi-dimensional Python lists using regular expressions. The process was simplified due to these intermediate text files having variables encapsulated in a mock-HTML format in anticipation of this step. This method was used rather than moving data directly from memory into tables during testing to avoid data loss in the event of system errors part-way into the testing process. Such measures were deemed necessary due to the extent of testing performed; in total, the runtime required to test all questions on all ten LLMs exceeded 16 hours. This was partly caused by an enforced 5-second delay with continuous readings of less than 10\% total system-wide CPU load, preventing damage to hardware, and ensuring consistency of resource usage readings.

Intermediate results were loaded into Pandas DataFrames, allowing specific columns to be extracted from the complete dataset, and exported into independent tables according to the tested category (i.e. mathematics, coding, etc.), in second and third normal form. Aggregate functions were applied to many tables to achieve practical data formats for analysis. From there, pandas.to\_excel() was called to export sub-tables into Excel files, where they could be formatted into a final state for visualization (see section~\ref{sec:results}).

\section{Results and Discussion} \label{sec:results}

\subsection{Correctness} \label{subsec:correctness}
This section contains charts that highlight key data obtained during testing. The figures of each subsection herein communicate the number of correct answers of all models combined for each test category. For each category, one or more charts presenting the individual question-solving performances of each LLM is provided. A summary of pertinent findings follows in section~\ref{subsec:resultsSummary}. Then, in section~\ref{subsec:resourceUsageResults}, mean resource utilization and execution times for each LLM across all tests are analyzed.

\subsubsection{Mathematics Test Results} \label{subsec:mathResults}

\begin{figure} [ht]
    \centering
    \includegraphics[width=1\linewidth]{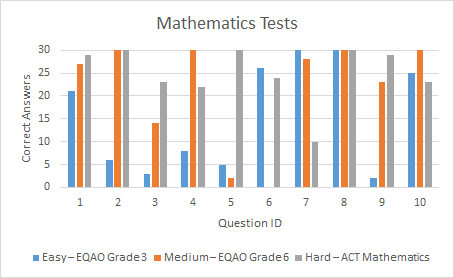}
    \caption{Mean score across all LLMs.}
    \label{fig:math_questions}
\end{figure}

\begin{figure} [ht]
    \centering
    \includegraphics[width=1\linewidth]{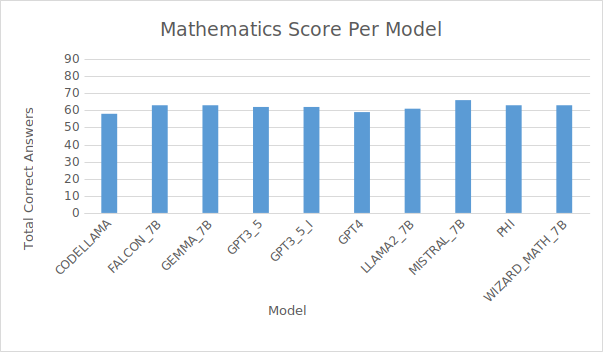}
    \caption{Total correct answers across all rounds.}
    \label{fig:math_allround}
\end{figure}

Results in this section were low on average as shown in Fig.~\ref{fig:math_questions}. Paradoxically, performance scaled inversely with difficulty level. A plausible explanation could be statistical disproportionality due to a different number of multiple-choice options per question, resulting in a higher number of lucky guesses, but easy and medium difficulties had four options each (i.e. A, B, C, D), while hard difficulty had five (i.e. A, B, C, D, E). Reference to the tested questions for each difficulty confirms that the level of challenge, from a human perspective, scales as initially expected, given the sources of each question set. It is possible that the models tested were more capable of interpreting the wording used in higher difficulty levels. Fig.~\ref{fig:math_allround} shows that Mistral 7B was the top performer in this category, with Code Llama 7B performing worst. Interestingly, GPT-4 was the second-worst performing model in this category. AI tutoring systems seeking to optimize performance when dealing with mathematics problems will garner no significant advantage from using proprietary models over self-hosted LLMs, since at present all LLMs seem to have relatively equal aptitude in this domain.

\subsubsection{Reading Test Results} 
\label{subsec:readingResults}

\begin{figure} [ht]
    \centering
    \includegraphics[width=1\linewidth]{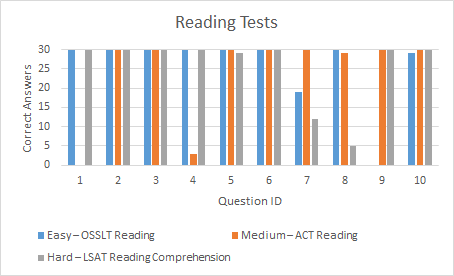}
    \caption{Mean score across all LLMs.}
    \label{fig:reading_questions}
\end{figure}

\begin{figure}[ht]
    \centering
    \includegraphics[width=1\linewidth]{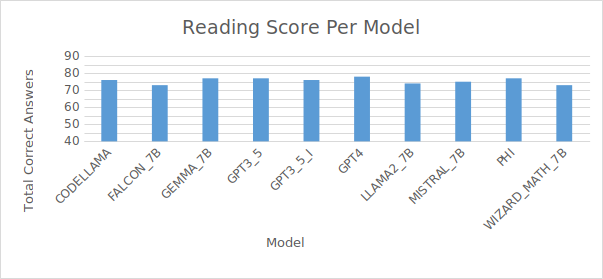}
    \caption{Total correct answers across all rounds.}
    \label{fig:reading_allround}
\end{figure}

Fig.~\ref{fig:reading_allround} shows the average results in this section were very strong. Falcon 7B and WizardMath 7B tied for last place, while GPT-4 had the highest ranking score. Fig.~\ref{fig:reading_questions} reveals that only a few questions were answered incorrectly; most notably, easy question 9, medium 1 and 4, and hard 8. These results demonstrate that all models provided consistent answers to any given question, but struggled significantly with the wording and style of some problems. ALMSs may benefit from modifying system prompts to achieve more consistent results when encountering questions whose wording style tends to be challenging for generic configurations like those used in these tests.

\subsubsection{Writing Test Results} \label{subsec:writingResults}

\begin{figure} [ht]
    \centering
    \includegraphics[width=1\linewidth]{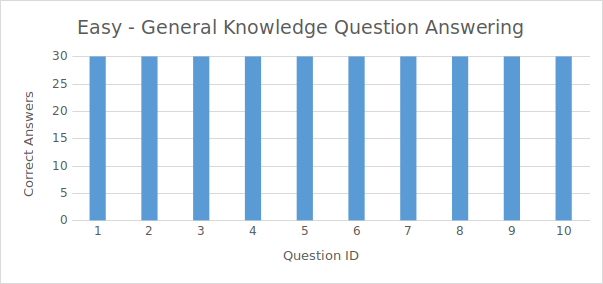}
    \caption{Mean score across all LLMs.}
    \label{fig:writing_easy}
\end{figure}

\begin{figure} [ht]
    \centering
    \includegraphics[width=1\linewidth]{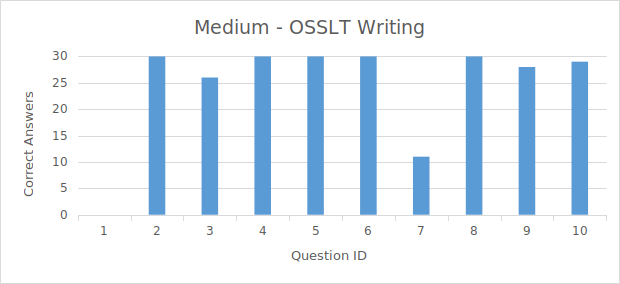}
    \caption{Mean score across all LLMs.}
    \label{fig:writing_medium}
\end{figure}

A summary chart detailing writing scores per model was not deemed relevant for this section, since inconsistent question types between difficulties made direct comparisons irrational. Overall, results were quite strong in this category. All models had perfect scores on the easy test, and very high scores on the medium test, excluding questions 1 and 7 as shown in Fig.~\ref{fig:writing_easy} and Fig.~\ref{fig:writing_medium}. Perfect scores across all of the easy test questions can be explained by the simplicity of the short-answer exercises, and the relaxed pass/fail grading scheme (see Section~\ref{subsec:writingTestDesign}). LLMs excel at low-complexity NLP writing tasks and general knowledge question answering, so it was expected that all models would perform well here. However, it was surprising to note the consistency in wording throughout responses to the easy-level questions, including between models. Despite models having no knowledge of their competition's responses and having no special input that would manipulate their output in this way, many models used the exact same, or similar words and phrases in response to each question. For example, question 3 directs the models to ``Write a haiku on honey bees." A correct response need only use a 5/7/5 syllable count, as per the definition of a haiku, and stay on topic. However, 7 out of 10 models responded by using the phrases ``Busy bees buzz by" or ``Busy bees buzzing" as the first clause, and 5 out of 10 models used the exact phrase ``Nature's sweet treasure" as the third clause. Similar trends were noted across other questions in this category. This suggests that many of these models were fine-tuned on the output of one or more other models, perhaps in an attempt to capitalize on their pre-existing successful tendencies. These findings are worthy of further investigation. On a practical note, given how often the same words and phrases were used in response to a given writing prompt, LLM use for academic dishonesty on short-answer questions could be easy to identify. Despite all models performing very well on the medium test, every LLM produced anti-perfect scores on question 1. All models unanimously chose multiple choice option ``C", while the correct answer was "B". A review of the problem suggests that its unique style and phrasing made option "B" stand out as a red herring, although the question's logic appears relatively simple for humans to understand.

\begin{figure} [ht]
    \centering
    \includegraphics[width=1\linewidth]{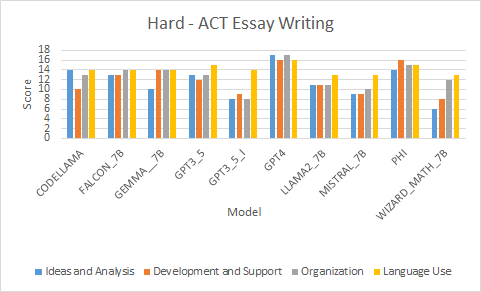}
    \caption{Scores per each grading metric.}
    \label{fig:writing_hard}
\end{figure}

\begin{figure} [ht]
    \centering
    \includegraphics[width=1\linewidth]{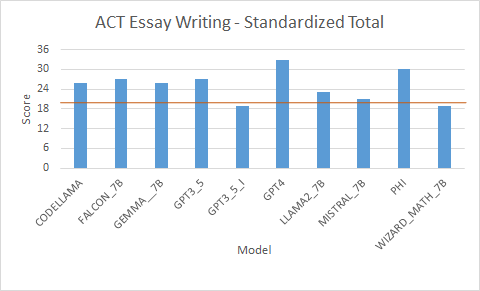}
    \caption{Total scores on the 36-point ACT grading scale. A score of 20 or more represents a passing grade.}
    \label{fig:writing_act}
\end{figure}

Fig.~\ref{fig:writing_hard} provides a breakdown of scores across each grading metric used, for each LLM. Trends observed here suggest that most models were especially strong in language use and generally weaker in ideas and analysis in comparison to other grading metrics. Fig.~\ref{fig:writing_act} shows the total essay score for each LLM, scaled to conform to the standardized 36-point ACT grading scale. The line at y=20 represents the cutoff score on the College Readiness Benchmark scale; models scoring at or above this value were deemed to perform at a level comparable to humans and thus achieved a passing grade for the purposes of this study. All models excluding GPT-3.5 Instruct and WizardMath 7B surpassed the cutoff grade. GPT-4 had the highest score at 33 points, with Phi-2 close behind at 30 points. See Section~\ref{subsec:writingTestDesign} for more details on the grading scheme used. These results show that in general an ALMS can rely on both proprietary and self-hosted LLMs to competently assist in tasks related to academic writing, although some consideration should be taken in deciding which models to delegate such tasks to.

\subsubsection{Reasoning Test Results} \label{subsec:reasoningResults}

\begin{figure} [ht]
    \centering
    \includegraphics[width=1\linewidth]{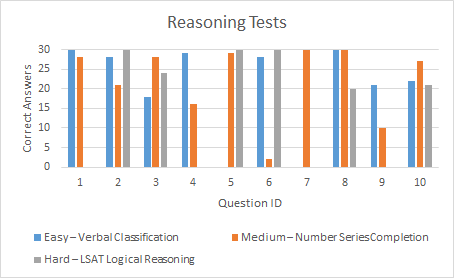}
    \caption{Mean score across all LLMs.}
    \label{fig:reasoning_questions}
\end{figure}

\begin{figure} [ht]
    \centering
    \includegraphics[width=1\linewidth]{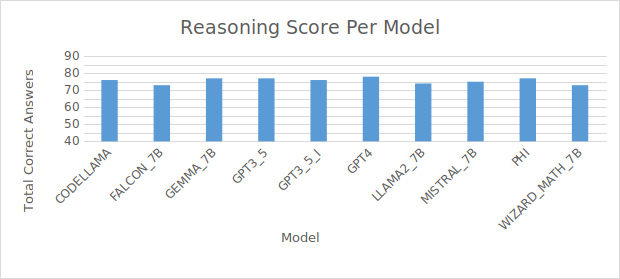}
    \caption{Total correct answers across all rounds.}
    \label{fig:reasoning_allround}
\end{figure}

Scores in this category were somewhat inconsistent, with a high degree of uniformity per each question, but also high variability in success rate throughout each difficulty. Given the challenging nature of the hard-level test, it was unsurprising that average scores were low as shown in Fig.~\ref{fig:reasoning_questions}. However, Fig.~\ref{fig:reasoning_allround} shows that for questions that were answered correctly even once, most models performed consistently well. Tuning an ALMS to approach this type of problem may prove difficult given the sporadic results noted here. 

\subsubsection{Coding Test Results} \label{subsec:codingResults}

\begin{figure}[ht]
    \centering
    \includegraphics[width=1\linewidth]{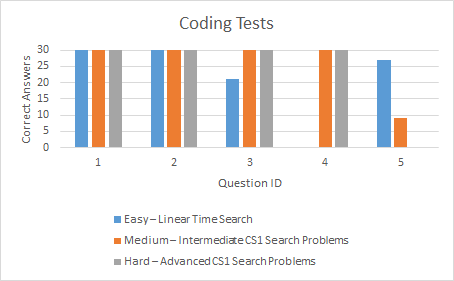}
    \caption{Mean score across all LLMs.}
    \label{fig:coding_questions}
\end{figure}

\begin{figure}[ht]
    \centering
    \includegraphics[width=1\linewidth]{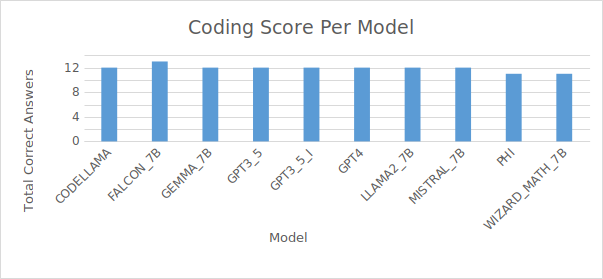}
    \caption{Total correct answers for coding across all rounds.}
    \label{fig:codingPerModel}
\end{figure}

Results from this section, summarized in Fig.~\ref{fig:codingPerModel}, indicate that the LLMs tested are more than capable of solving basic search problems at a skill level reflective of students in post-secondary CS1 courses. Falcon 7B scored the highest, but only by a single point. Detailed results across each difficulty and between individual questions are presented in Fig.~\ref{fig:coding_questions}. Most LLMs tested performed poorly on easy question 4, medium 5, and hard 5, but unanimously succeeded on the majority of other problems. ALMSs that endeavour to assist learners with basic coding problems are likely to find success, and may gain a slight advantage with careful selection of the appropriate LLM.
 
\subsubsection{Section Summary} \label{subsec:resultsSummary}
Overall, we find that compared to statistics of human subjects who have taken similar tests, the performances of models benchmarked in this study were:

\begin{itemize}
    \item Weak in mathematics
    \item Strong in reading
    \item Strong in writing
    \item Moderate in reasoning
    \item Strong in coding
\end{itemize}
    
Notably, models indicated a high degree of uniformity in their response selection; most often, they did either very well or very poorly on a given question. This suggests that they are greatly influenced by the wording of each problem, but that their underlying problem-solving ability is relatively high. Also interesting was the high performance of self-hosted models. These models had similar scores to proprietary models with exponentially higher parameter counts in most categories. While there was a disproportionate distribution between GPT-4 and other models in the hard-level writing test, Phi-2, the lowest parameter model, achieved scores close to GPT-4 and outpaced all other models in this exercise. An ALMS might optimize its performance by delegating different problem types to the most suitable LLM, for example Phi for reading and writing tasks, and Mistral for mathematics. In this way a small collection of self-hosted models may rival considerably larger proprietary LLMs in performance, while retaining the other advantages of self-hosting like data privacy and low-cost operation.

\subsection{Resource Utilization Test Results} \label{subsec:resourceUsageResults}
\begin{figure}[ht]
    \centering
    \includegraphics[width=1\linewidth]{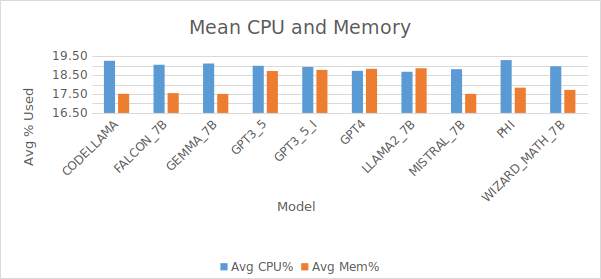}
    \caption{Mean resource usage across all tests.}
    \label{fig:resource_usage}
\end{figure}

\begin{figure}[ht]
    \centering
    \includegraphics[width=1\linewidth]{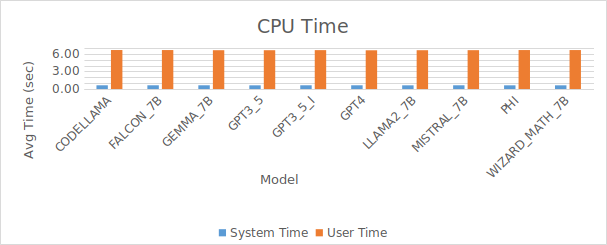}
    \caption{Mean CPU time across all tests.}
    \label{fig:cpu_time}
\end{figure}

\begin{figure}[ht]
    \centering
    \includegraphics[width=1\linewidth]{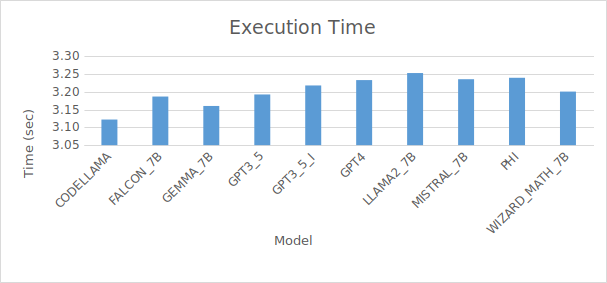}
    \caption{Mean execution time across all tests.}
    \label{fig:execution_time}
\end{figure}

This section presents the mean system resource consumption and execution times across all questions, for each LLM. Fig.~\ref{fig:resource_usage} displays average percentages of CPU and virtual memory used; Fig.~\ref{fig:cpu_time} translates CPU time (the duration the application spent occupying the CPU) between system time and user time; and Fig.~\ref{fig:execution_time} shows the time each LLM spent solving each question on average.

The results of the resource utilization tests were very consistent across all models and metrics, with a few exceptions. Mean memory usage was slightly higher on average for all three proprietary models compared to mean readings across self-hosted models. This was surprising since the expectation was that self-hosted models would be consistently more resource-intensive than those accessed via API calls. Llama2 7B returned average memory usage readings on par with the proprietary models; this, combined with its mean execution time being the highest of all models tested, makes it the lowest-performing model in terms of hardware utilization. However, the range between the highest and lowest-performing models was so small that this was not a significant concern. Of greater interest was the fact that across all hardware usage metrics, including execution time, the proprietary models performed at an equal or slightly inferior level to self-hosted models. In the context of ALMS design, these results indicate that there is a negligible difference in resource usage and execution time between models, assuming that the parameter size of self-hosted LLMs is suitable for the device's hardware specifications.

\subsection{Confounding Variables} \label{subsec:confoundingVars}
Based on the 3,780 total questions answered during testing, 35 did not include an identifier (i.e. A, B, C, D, or E). Upon analyzing these specific responses, it was discovered that 12 claimed that there was ``no valid option" and 20 provided an answer, but failed to include the corresponding identifier, as they were instructed to. This data is represented in Table~\ref{tab:confoundingVars}. Since this is such a small portion of the overall sample size, it should not have any meaningful effect on the study's results.

\begin{table}[ht]
    \centering
    \caption{Responses without Identifiers}
    \label{tab:confoundingVars}
    \begin{tabularx}{\columnwidth}{c|c|c}
        \hline
         ``No valid option" & Explanation Only & Total Void Responses \\
         \hline
         12 & 20 & 35 \\
         \hline
    \end{tabularx}

\end{table}

\section{Conclusion and Future Work} \label{sec:conclusionAndFuture}
An adaptive learning management system may incorporate many different technologies to varying effect. Expert systems offer more precision and speed of execution but take longer to implement. An LLM-based approach is much faster to configure and has access to unique NLP features. However, self-hosted LLMs are more hardware intensive than ESs and include a delay when processing results, with LLMs accessed via API calls carrying additional financial costs. Issues with LLMs may be partially mitigated by weighing the hardware requirements of self-hosted options against the cost and increased performance of API calls to proprietary models, and through the use of RAG, vector embedding, and prompt engineering. Self-hosted LLMs also diminish privacy risks associated with using proprietary models, given the option to perform all execution client-side. A hybrid approach may be the most practical option, retaining the strengths of both ES and LLM implementations while circumventing their shortcomings. Through the extensive testing performed in Phase III of the project, we found that self-hosted LLMs were very capable, both in terms of problem-solving competency and hardware utilization, relative to proprietary models accessed via API. This testing also revealed that in comparison to statistics of human subjects who have attempted similar tests, modern LLMs demonstrate strong reading, writing, and CS1-level coding abilities, and moderate reasoning performance, but tend to struggle in mathematics. Considering that research in the advancement of low-parameter LLMs is ongoing, we speculate that self-hosted options will become even more viable soon.

The ALMS prototype developed in this project shows promise, but possibilities gleaned throughout the development process may be applicable to future projects. Phase I's prototype was successful in using a manually configured expert system with a custom knowledge base to relay information to the user. Our implementation used a rigid UI as a presentation medium. It may be possible to build a human-curated knowledge base as the primary information source for an LLM-driven system. This could be accomplished by converting the knowledge base into a \textit{vectorstore} using vector embedding, and then directing the LLM to draw from the store using RAG whenever learning resources are requested by the system. Findings in Phase II demonstrated the value of RAG for augmenting LLM responses by limiting hallucinations and allowing citation of sources. Investigating the possibilities of local \textit{vectorstores} that retain vector output from embedding models on local storage devices between queries could greatly speed up RAG responses that repeatedly draw from the same source material. Solutions like \textit{Pinecone} and \textit{Chroma} appear promising. Phase III of the project highlighted the possible strengths and weaknesses of a selection of LLMs. Further investigation into why models consistently struggled with the wording of specific questions, despite being capable of solving similar questions of equal or greater difficulty, could illuminate a fundamental issue with current NLP technology. Research into the cause behind consistently identical or extremely similar responses produced by different LLMs during the easy-difficulty writing questions may reveal that some models are fine-tuned on the output of others. Related studies questioning the long-term value of models that rely on fine-tuning over other types of training would aid ALMS developers in the selection of suitable LLMs. Continued benchmark testing covering a wider breadth of questions, analyzing new models as they are released and updated could provide insights into the aptitude of the current state-of-the-art LLMs relative to human baselines. Other LLM benchmarks do exist, but measuring LLM performance in comparison to student averages on standardized tests is especially relevant for understanding the potential of LLMs as a component in an ALMS.

\section*{Acknowledgment}  
Special thanks to Jennifer Hopkins, Steve Sheehan, Vanessa Caddel, and Ryan Langlois for assistance in grading the AI-composed ACT essays and providing data for the hard-level writing test in section~\ref{subsec:writingResults}. Partial expenses for this project were supported by Laurentian University Grant Number: 10-1-1092634.

\bibliographystyle{IEEEtran}        
\bibliography{bibtex/bib/reference}           
\ifCLASSOPTIONcaptionsoff
  \newpage
\fi

\end{document}